\documentclass{article}
\usepackage{lettrine}

\PassOptionsToPackage{square,numbers}{natbib}
\usepackage[final]{nips_2016}

\usepackage[utf8]{inputenc} 
\usepackage[T1]{fontenc}    
\usepackage{hyperref}       
\usepackage{url}            
 \usepackage{booktabs}       
\usepackage{amsfonts}       
\usepackage{nicefrac}       
\usepackage{microtype}      
\usepackage{subfigure}
\usepackage{multirow}
\usepackage{color}
\usepackage{xcolor}
\usepackage{array}
\usepackage{graphicx}
\usepackage{amsmath}
\usepackage{algorithm}
\usepackage[noend]{algpseudocode}
\usepackage{afterpage}

\makeatletter
\newcommand\fs@nobottomruled{\def\@fs@cfont{\bfseries}\let\@fs@capt\floatc@ruled
  \def\@fs@pre{\hrule height.8pt depth0pt \kern2pt}%
  \def\@fs@post{}
  \def\@fs@mid{\kern2pt\hrule\kern2pt}%
  \let\@fs@iftopcapt\iftrue}
\makeatother

\floatstyle{nobottomruled}
\restylefloat{algorithm}

\definecolor{yellow}{RGB}{255,255,0}

\newsavebox{\fmbox}
\newenvironment{fmpage}[1]
{\begin{lrbox}{\fmbox}\begin{minipage}{#1}}
{\end{minipage}\end{lrbox}\fbox{\usebox{\fmbox}}}

\title{Deep Learning for Explicitly Modeling \\Optimization Landscapes}

\author{
  Shumeet Baluja \\
  Google, Inc.\\
  Mountain View, CA. 94043\\
  \texttt{shumeet@google.com} \\
}

\begin{document}

\maketitle

\begin{abstract}

  In all but the most trivial optimization problems, the structure of
  the solutions exhibit complex interdependencies between the input
  parameters.  Decades of research with stochastic search techniques
  has shown the benefit of explicitly modeling the interactions
  between sets of parameters and the overall quality of the solutions
  discovered.  We demonstrate a novel method, based on learning deep
  networks, to model the global landscapes of optimization problems.
  To represent the search space concisely and accurately, the deep
  networks must encode information about the underlying parameter
  interactions and their contributions to the quality of the solution.
  Once the networks are trained, the networks are probed to reveal
  parameter combinations with high expected performance with respect
  to the optimization task.  These estimates are used to initialize
  fast, randomized, local search algorithms, which in turn expose more
  information about the search space that is subsequently used to
  refine the models.  We demonstrate the technique on multiple
  optimization problems that have arisen in a variety of real-world
  domains, including: packing, graphics, job scheduling, layout and
  compression.  The problems include combinatoric search spaces,
  discontinuous and highly non-linear spaces, and span binary,
  higher-cardinality discrete, as well as continuous parameters.
  Strengths, limitations, and extensions of the approach are
  extensively discussed and demonstrated.

\end{abstract}

\section{Introduction to Optimization via Search Space Modeling}
\label {intro}

In the 1990s, a number of
researchers~\cite{baluja1994population}~\cite{juels1996topics}~\cite{muhlenbein1996recombination}~\cite{harik1999compact}
independently started employing probabilistic models to guide
heuristic stochastic based-search algorithms.  The idea was simple, to
use the knowledge of the search landscape that may be ascertained by
analyzing the points encountered to guide where to look next.  This
sharply contrasted the procedure of many successful randomized
hill-climbing algorithms that made small perturbations stochastically
in hopes that a better solution was a close neighbor to the current
best solution.

In the simplest instantiation, probabilistic methods explicitly
maintain statistics about the search space by creating models of the
good solutions found so far. These models are then sampled to generate
the next query points to be evaluated. The sampled solutions are then
used to update the model, and the cycle is continued.  Probabilistic
models for optimization were motivated with three goals: as a new
optimization method, as a method to incorporate simple learning
(Hebbian style probability updates) into hillclimbing, and as a method
to explain how genetic algorithms
~\cite{goldberg1989genetic}~\cite{holland1975adaptation}~\cite{de2005genetic}
might work.

A bit more detail on the third motivation is warranted: by maintaining
a population of points (rather than a single point from which search
is centered), genetic algorithms (GAs) can be viewed as creating
\emph{implicit} probabilistic models of the solutions seen in the
search. GAs attempt to implicitly capture dependencies between
parameters and solution quality through the distribution of parameter
settings contained in the population of solutions.  Exploration
proceeds by generating new samples to evaluate through a process of
applying randomized ``recombination/crossover'' operators to pairs of
high-performance members of the population.  Because only the
high-performance are selected for recombination, parameter settings
found in the poor-performing solutions are not explored further.  The
recombination operator, which combines two ``parents'' candidate
solutions by producing ``children'' that transfer sets of parameters
from each parent into the children, serves to keep sets of parameters
together (though in a randomized manner) in the next set of candidate
solutions that are explored.  In terms of sampling, this can be viewed
as a simple approach to sampling the population's statistics.  Though
similar in intent, this implicit manner of sampling the distribution
of solutions contrasts with the approach presented in this paper.
Here, \emph{explicit} steps are taken to model the parameters
interdependencies that contribute to the quality of candidate
solutions.  The goal of this paper is succinctly stated as follows:

\begin{center}
\begin{fmpage}{5in}

\lettrine[findent=1pt]{\textbf{T}}{}he primary goal of this paper is
to demonstrate that while employing any search algorithm
(hillclimbing, GA, simulated annealing, etc.), it is possible to
simultaneously learn a deep-neural network based approximation of the
evaluation function.  Then, through the use of ``deep-network
inversion'' (employed as a method to sample the deep networks),
intelligent perturbations of some/all of the samples generated by the
search algorithm can be made.  After these perturbations, the the new
solutions should have an increased probability of higher scores. ~\\

As a secondary benefit, this modeling may also be used to make the
evaluation more efficient by replacing it with the DNN's approximation
of the evaluation function.

\end{fmpage}
\end{center}

\subsection {Predecessors to Deep Modeling of Optimization Landscapes}

Within the GA literature, one of the first attempts at using population
level statistics was the “Bit-Based Simulated Crossover (BSC)”
operator~\cite{syswerda1993simulated}. Instead of combining pairs of
solutions, population-level statistics were used to generate new
solutions. The BSC operator worked as follows. For each bit position,
the number of population members which contain a one in that bit
position were counted. Each member’s contribution was weighted by its
fitness with respect to the target optimization function. The same
process was used to count the number of zeros. Instead of using
traditional crossover operators to generate new solutions, BSC
generated new query points by stochastically assigning each bit’s
value by the probability of having seen that value in the previous
population (the value specified by the weighted count).  The important
point to note is that BSC used the population statistics to generate
new solutions.

Extending the above idea and incorporating Hebbian learning, another
early probabilistic optimization was the Population-Based Incremental
Learning algorithm (PBIL)~\cite{baluja1994population}.  Rather than
being based on population-genetics, the lineage for PBIL stems back to
early reinforcement learning. PBIL is akin to a cooperative system of
discrete learning automata in which the automata choose their actions
independently, but all automata receive a common reinforcement
dependent upon all their actions~\cite{thathachar1987learning}. Unlike
most previous studies of learning automata, which have commonly
addressed optimization in noisy but very small environments, PBIL was
used to explore large deterministic spaces.

The PBIL algorithm, most easily described with a binary alphabet,
works as follows.  The algorithm maintains a real-valued probability
vector, specifying the probability of generating a 1 or a 0 in each
bit position (just the first order statistics).  The probability
vector is sampled repeatedly to generate a set of new candidate
solutions.  These candidates are evaluated with respect to the
optimization function, and the best solution is kept and all the other
discarded.  The probability vector is ``moved'' towards the best
solution through simple Hebbian-like updates.  And the cycle is
repeated.  Note that the probabilistic model created in PBIL is
extremely simple: \emph{There are no inter-parameter dependencies
  captured; each bit is modeled independently. The entire probability
  model is a single vector}.  Although this simple probabilistic model
was used, PBIL was successful when compared to a variety of standard
genetic algorithm and hillclimbing procedures on numerous benchmark
and real-world problems. A more theoretical analysis of PBIL can be
found
in~\cite{juels1996topics}\cite{kvasnicka1995hill}\cite{Hohfeld1997}\cite{rastegar2005convergence}\cite{gonzalez2001convergence}\cite{lozano2000analyzing}.

The most immediate improvements to PBIL are mechanisms that capture
inter-parameter dependencies.  Mutual Information Maximization for
Input Clustering (MIMIC)~\cite{de1997mimic} was one of the first to do
this. MIMIC captured a heuristically chosen set of the pairwise
dependencies between the solution parameters.  From the top N\% of all
previously generated solutions, pair-wise conditional probabilities
were calculated. MIMIC then used a greedy search to generate a chain
in which each variable was conditioned on the previous variable. The
first variable in the chain, $X_1$, was chosen to be the variable with
the lowest unconditional entropy, $H(X_1)$. When deciding which
subsequent variable, $X_{i+1}$, to add to the chain, MIMIC selected
the variable with the lowest conditional entropy, $H(X_{i+1} | X_i )$.
This extended the solely-unconditional model with PBIL to maintain a
set of pair-wise dependencies.  In~\cite{baluja1998fast}, MIMIC’s
probabilistic model was extended to a larger class of dependency
graphs: trees in which each variable is conditioned on at most one
parent.  As shown in 1968, this created the optimal tree-shaped
network for a maximum-likelihood model of the data
~\cite{chow1968approximating}. In experimental comparisons, MIMIC’s
chain-based probabilistic models typically performed significantly
better than PBIL’s simpler models. The tree-based graphs performed
significantly better than MIMIC’s chains.

The trend indicated that more accurate probabilistic models increased
the probability of generating new candidate solutions in promising
regions of the search space~\cite{baluja1997using}.  The natural
extension to pair-wise modeling is modeling arbitrary dependencies.
Bayesian networks are a popular method for efficiently representing
dependencies~\cite{heckerman2008tutorial}~\cite{pearl2000bayesian}.
Bayesian networks are directed acyclic graphs in which each variable
is represented by a vertex, and dependencies between variables are
encoded as edges.  Numerous researchers have taken the step of
combining full Bayesian networks with stochastic search
~\cite{pelikan1999boa}~\cite{yao2015bayesian}~\cite{etxeberria1999global}.
For an overview, see ~\cite{pelikan2007scalable}.

An alternate model building approach, termed STAGE, was presented
in~\cite{boyan2000learning}.  STAGE attempts to map a set of
user-supplied features of the state space to a single value. This
value represents the quality of solutions found thus far.  The “value
function” is used to select the next point from which to initialize
search.  Unlike the other algorithms described above, which attempt to
automatically model the effects of parameter combinations on the
overall solution quality, here hand-created features related to the
optimization function are used.  Other than the use of hand-crafted
features, this approaches shares many of the important characteristics
of the methods used in this paper.

In the next section, we describe the Deep-Opt algorithm and give
details of how the probabilistic model is created and sampled.  We
also describe how the model is integrated with fast-search heuristics,
following the work of~\cite{boyan2000learning}
and~\cite{baluja1998fast}.  In Section~\ref{tests}, we examine the
performance of Deep-Opt on a wide variety of test problems.  There are
many interesting alternatives and extensions to the approach used
here; experiments with three are presented in
Section~\ref{alternatives}.  The three alternatives help to increase
the set of problems that can be tackled and address some of the limits
of the approach as described in Section~\ref{algorithm}.  Finally, a
discussion of the results and suggestions for future work are
presented in Section~\ref{conclusions}.

\section{Deep Learning for Search Space Modeling}
\label{algorithm}

Optimization with probabilistic modeling, at a high level, is simply
explained in Figure~\ref{fig:highLevel}. As shown in the figure, in
previous work, we started with a large set of candidate solutions and
evaluated them with respect to the objective function of the problem
to be solved (\emph{e.g.} the total tour length of the classic
traveling salesman problem).  The poor-performing candidate solutions
were discarded.  The set of remaining solutions, usually a small
subset of the better performing members from the original set, were
then modeled.  The model was stochastically sampled, the solutions
evaluated, and the process was continued.

\begin{figure}[h]
 \centering
 \begin{fmpage}{\textwidth}
  \begin{algorithm}[H]
  \caption{High Level Probabilistic Modeling for Optimization}      
  \begin{algorithmic}
    \\
    \\
  \State Create set, S, with random solutions from uniform distribution.
  \While {termination condition is not met}
  \State Create a probabilistic model, PM, of S.
  \State Stochastically sample from PM, to generate C candidate solutions.
  \State Evaluate the C candidate solutions.  
  \State Update S with  \emph{only the N high-evaluation} solutions
  from $C$, where $N \ll C$.
  \EndWhile
  \\
  \end{algorithmic}
\end{algorithm}

\end{fmpage}  
  \caption{Probabilistic Modeling for Optimization Overview.}
  \label{fig:highLevel} 
  \vspace {0.1in}
\end{figure}

In the interest of being concrete, a simple example of a probabilistic
model is provided in Figure~\ref{fig:popModel} using PBIL's
independent-parameter model to represent 4 different
populations \footnote {Note that in a variety of previous studies, the
  solution-vectors were represented as a string of binary parameters
  as this worked well with the operators used with evolutionary
  algorithms.}.  Once the model, $PM$, is created, $PM$ is sampled to
generate the candidate solutions to evaluate next.  Why does this idea
work?  The solutions that are represented in $PM$ are only those with
high evaluations (the poor performing ones were discarded and never
modeled), therefore, those that are subsequently generated by sampling
the statistics of multiple good solutions should be high-performing as
well.  A large number of solutions are generated, each is evaluated,
and the lower-performing ones discarded, and the procedure is
repeated~\footnote{Note that in the $PM$ sampling process, a small
  amount of random mutation/perturbations are also introduced into the
  generated solutions to ensure that a heterogeneous set of solutions
  are explored.}.

\begin{figure}
  \vspace {0.1in}
  \centering
  \label{fig:popModel}
  \small
  \begin{fmpage}{0.22\textwidth}
    Population A:\\
    \\
    \texttt{
    1000\\
    0010\\
    1000\\
    1001\\}
    ---\\
    0.75, 0.00, 0.25, 0.25\\
  \end {fmpage}
  \begin{fmpage}{0.22\textwidth}
    Population B:\\
    \\
    \texttt{    
    1010\\
    1010\\    
    0101\\
    0101\\}
    ---\\    
    0.50, 0.50 0.50, 0.50\\
  \end {fmpage}
  \begin{fmpage}{0.22\textwidth}
    Population C:\\
    \\
    \texttt{    
    1000\\
    0010\\    
    0000\\
    0000\\}
    ---\\    
    0.25,0.00,0.25,0.00\\
  \end {fmpage}
  \begin{fmpage}{0.22\textwidth}
    Population D:\\
    \\
    \texttt{    
    1111\\
    0000\\    
    1111\\
    0000\\}
    ---\\
    0.50, 0.50 0.50, 0.50\\
  \end {fmpage}

   \caption{Modeling 4 different binary-populations of 4 samples each.
     In PBIL's probabilistic model, each bit position is modeled
     independently and is simply the probability of having a '1' in
     each bit position.  This distribution can easily be sampled to
     generate new points to evaluate.}
   \vspace {0.1in}     
\end{figure}

Note that although PBIL's probabilistic model is extremely simple, and
no inter-parameter dependencies are modeled, as mentioned earlier,
even this proved effective in many standard optimization tasks.
Despite the successes, however, as shown in Figure~\ref{fig:popModel}
there are severe limitations: population sets B \& D, although
different, are represented with the same probabilistic vector; thereby
demonstrating the inadequacy of the models and the need for more
powerful representations of the candidate solution's statistics.
Thus, models such as MIMIC \& Optimal-Dependency-Trees (described in
Section~\ref{intro}), in which pair-wise or higher-order dependencies
were represented, were introduced.  As problem complexity increased,
these more flexible and powerful models improved the quality of the
solutions generated.

In this paper, we use a neural network to create a mapping between the
solutions sampled to that point and their score, as determined by the
evaluation function.  In the context of describing the algorithm, we
also expose four of the largest differences between our approach and
the approaches used earlier.

\paragraph {(1)} One of the primary differences is that by using a deep neural network,
we do not \emph{a priori} specify the form of the dependencies in the
probabilistic model (\emph{e.g.} pair-wise, or triplet combinations,
etc.)  Although the architecture of the neural network used for
modeling is manually specified, the actual dependencies that the
network encodes need not be the same form for all parameters, nor are
they deeply tied to architecture of the
network~\cite{hornik1989multilayer}.  For all but one of the tests
described in this paper, the neural network architecture will not
change; a deep network is capable of modeling the necessary
statistics.

\paragraph {(2)} Once the network has learned the mapping from input parameters
to evaluation, a procedure for \emph{generating} the new points to evaluate
is necessary.  This procedure is quite different than the
straight-forward sampling that was possible in the previous models
such as those in PBIL or dependency trees.  In the simplest model,
PBIL (see Figure~\ref{fig:popModel}), samples were easily generated by
a biased-random sample, where a '1' was generated in position $p$
independently from any other bit, as specified by the real-value in
position $p$ in the probability vector.  With models with
dependencies, such as the
dependency-trees~\cite{chow1968approximating}, the sampling is simply
conditioned on the variables on which the parameter is dependent (as
specified in the tree-based model).

With neural networks, however, generating samples is more complex.  It
is based on the technique of ``network inversion'': \textbf{given a
  trained network}, network inversion uses standard back-propagation
to ~\emph {modify the inputs} rather than the network's weights.  The
inputs are modified to match the preset and clamped outputs. This
method was first presented in~\cite{linden1989inversion} and has
recently been popularized within the context of texture and style
generation with neural networks
~\cite{gatys2015texture}~\cite{gatys2015neural}.

We use it as follows: We are given a trained network that maps the
input parameters (scaled between 0.0 and 1.0) to their evaluation
(also scaled between 0.0 and 1.0).  First, all the weights of the
network are frozen; they will not change for the sample generation
process.  Second, we clamp the output to the desired output --- for
maximization problems, we clamp the output 1.0; this indicates that we
would like to generate solutions that are as good as the best ones
seen so far.

The inputs are then initialized (either randomly or by other means
such as perturbing the best solution seen thus far) and the network
performs a forward propagation step and the error is measured at the
output.  The error metric is the standard Least-Mean Squares Error on
the target output (which is clamped to 1.0).

$$ E_{LMS} = \sum_{i \in outputs} (targets_i - predicted_i)^2 $$

Since we pin the target to 1.0, and there is only a single output
scaled between 0.0 and 1.0 (that represents the score of the candidate
solution).  This is simply:

$$ E_{LMS} =  (1.0 - predicted_{score})^2 $$

A process similar to standard training with stochastic-descent
back-propagation (or any other network training procedure) is then
used.  However, unlike standard training, the errors are propagated
all the way back to the inputs, and the \textbf{inputs are modified --
  not the weights}.  As described in~\cite{linden1989inversion}, the
procedure addresses the following question: ``Which input should be
fed into the net to produce an output which approximates the given
target vector T' (in our case the the target vector T is a simple
scalar of 1.0).  The error signal for the input $i$:

$$ \delta_i = - \dfrac{\delta E}{\delta input_i } $$

tells the \emph{input units} how to change (direction and magnitude)
to decrease the error~\footnote{Whenever the units are changed outside
  the bounds of [0.0,1.0], they are clipped to be within the range.
  An alternative approach could have passed the activations through
  sigmoid functions; however, this would make setting values at the
  extrema (0.0 and 1.0) slower.}.  In general, modest learning-rates for
  the gradient descent algorithm were found to work best for the
  network-inversion process (we used the Adam
  Optimizer~\cite{kingma2014adam} with learning-rate = 0.001).  If the
  networks are trained well such that they successfully model the
  search landscape, then the set of input values found through this
  procedure will yield high-evaluation solutions when tested on the
  actual evaluation function.

\paragraph {(3)}   Once the new candidate solutions are generated and
evaluated, what is the next step?  In the previous uses of
probabilistic models (Figure~\ref{fig:highLevel}), the low-performance
candidate solutions were discarded and the high-performance solutions
were kept.  Interestingly, the actual evaluation of the
high-performance solutions was not used.  In contrast, in our
procedure, we create an explicit mapping from the candidate solution
to its score~\footnote{In our previous explorations of probabilistic
  models, some instantiations had weighted the contribution of each
  member of $S$ by their relative score.  Although this changes the
  models by specifying how well each samples is represented in the
  model, it \emph{does not} create an explicit mapping from the
  parameters to their scores.}.  This reflects a fundamental difference
in approaches: since an explicit mapping is created, it is not assumed
that the model only represents good solutions; it represents both high
and low quality solutions.  

\paragraph {(4)}  Finally, note that in contrast to many previous
studies that represented the solution vectors as binary strings and
modeled only the binary parameters, the deep-neural networks used in
this study naturally model real-values.  Extensions to binary and
other discrete parameters will be described in
Section~\ref{alternatives}.

\begin{figure}
  \centering
 \begin{fmpage}{\textwidth}  
  \begin{algorithm}[H]
  \caption{Next-Ascent Stochastic Hillclimbing (NASH) (shown for maximization)}  
  \begin{algorithmic}
    \\
    \\
  \State Create a random candidate solution, $c$, composed of $|P|$ real-valued parameters.
  \State BestEvaluation $\gets$ Evaluate ($c$)
  \While {termination condition is not met}
    \State $c' \gets c$  
    \State Number-Perturbations = Select-Random-Integer from $[1 ,  m \times |c|]$
    \Loop {~Number-Perturbations}
        \State Randomly select a parameter, $p$, from $c'$
        \State c' $\gets$ Modify $p$ to a randomly chosen value between $[(0.75*p),(1.25 * p)]$
    \EndLoop
    
    \State CandidateEvaluation $\gets$ Evaluate (c')
    \If {(CandidateEvaluation $\ge$ BestEvaluation)}
       \State $c \gets c'$
       \State BestEvaluation $\gets$ CandidateEvaluation
    \Else
       \State Discard $c'$
    \EndIf
    \EndWhile
    \\
    Return c
    \\
  \end{algorithmic}
  \end{algorithm}
  \end{fmpage}

  \caption{Next-Ascent Stochastic Hillclimbing (NASH). In our
    implementation we set $m = 2\%$.  Note that the mutation amount
    $\pm 0.25$ is quite large; however, this was set based on
    extensive empirical testing on these and similar problems. |P| is
    dependent on the problem.  In the experiments presented throughout
    this paper, it ranged from 50 to 1600. }
  \label{fig:nash} 
   \vspace {0.1in}  
\end{figure}

\subsection {Integration with Fast Local Search Heuristics}

In the simplest implementation, the candidate solutions generated by
the network inversion are evaluated and the cycle is continued.
Although this method will work, there are drawbacks.  First, this is a
slow process; training a full network to map the sampled points to
their evaluations is an expensive procedure, as is sampling the
network.  Second, a post-processing step of local optimization, where
small changes are made to the solutions generated, yields improvement
to the solutions found.  This is because both the interpolation and
extrapolation capabilities of the trained networks are not perfect;
there will be discrepancies between the estimated ``goodness''
(evaluation) of a candidate solution and its actual evaluation.

Two early works in probabilistic model-based
optimization~\cite{baluja1998fast}~\cite{boyan2000learning} suggested
the use of the probabilistic models as methods to initialize faster
local-search optimization techniques.  This technique is used here.  A
very simple next-ascent stochastic hillclimbing (\emph{NASH}) procedure
is shown in Figure~\ref{fig:nash}.  It has repeatedly proven to work
well in practice when used in conjunction with other optimization
algorithms to perform local optimization, and also surprisingly well
when used alone in a variety of scenarios
~\cite{juels1996stochastic}~\cite{baluja1995empirical}~\cite{mitchell1994}~\cite{covellTraffic2015}.
For the majority of the paper, we will use NASH as the underlying
search process; the neural modeling will ``wrap-around'' NASH.

Importantly, note that NASH is initialized with a single candidate
solution.  This candidate is perturbed until an equal or better
solution is found.  This fits well into the procedures described thus
far: all of the candidate solutions evaluated by NASH can be added to
the pool of solutions to be modeled, $S$.  When the networks are
trained and a next set of candidate solutions are generated, the
single best solution from them is used to initialize the NASH
algorithm.  NASH proceeds as normal, again recording all the candidate
solutions it evaluates -- which are then use to augment $S$ in the
next time step, and the cycle continues.

\begin{figure}
  \centering
\includegraphics[width=\textwidth]{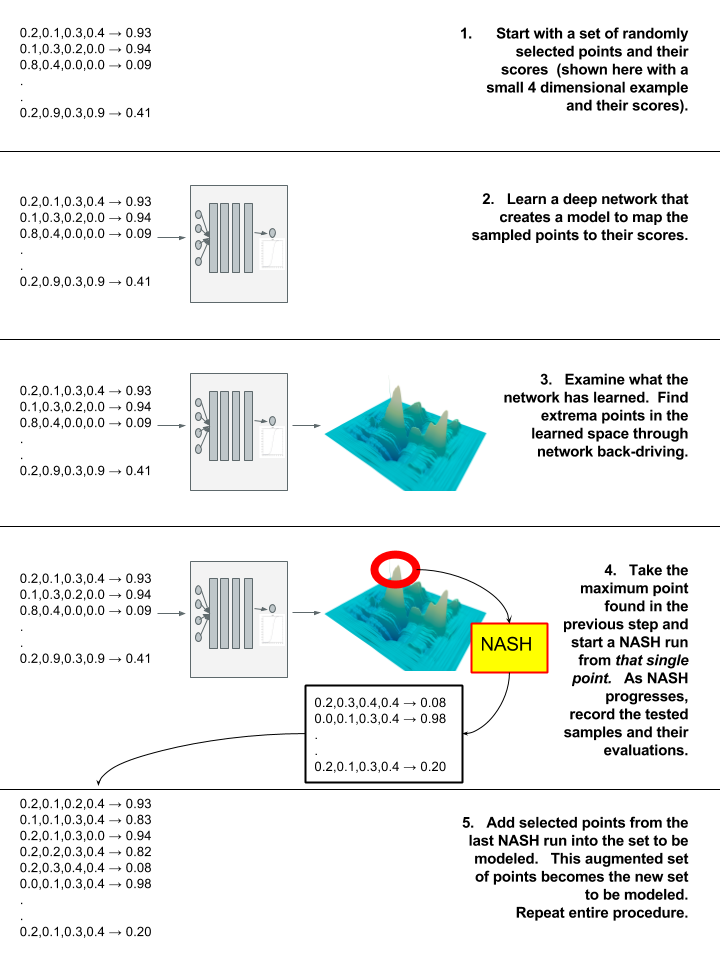}  
\caption{A visual description of Deep-Opt. \scriptsize{ (Sample topographic plot by~\cite{plotly}) } }
  \label{fig:steps}

\end{figure}

A visual description of the algorithm is shown in Figure~\ref{fig:steps}.

\subsection {Putting it all Together}

In this section, we describe some of the pragmatic considerations for
deployment, and put all of the steps into step-by-step directions.

There are many possible ways to create and maintain $S$, the samples
that are used for training the modeling networks.  In our study, the
size of S is kept constant throughout the run.  Although a
sufficiently large deep neural network should be able to represent the
surface represented by all of the points seen, it would require both
extra computation and not be valuable in practice.  Overly precise
models of low-evaluation areas are not necessary.  To keep $S$ a
constant size, after every NASH run, the last 1000 unique solutions
from the NASH run are added to $S$, and $S$ is pruned back by removing
the members that have been in S the longest (this implements a simple
first-in-first-out scheme), as was suggested in previous
studies~\cite{baluja1998fast}.  If the algorithm is progressing
correctly, the average score of the solutions present in $S$ increases
over time.

The full Deep-Opt algorithm is shown in Figure~\ref{fig:deepopt}.
With respect to the step ``Train a Deep Neural Network'', there are
several points that should be noted.  The use of a validation set is
vital to good performance.  In each training cycle, (100) samples are
drawn by perturbing the best solution and added to the validation set.
Then, after each step in training, the correlation between the actual
evaluation of these samples and the network's predicted evaluation of
these samples is measured.  When the correlation begins to decrease,
training is stopped and the next step of the algorithm begins.  If the
correlation is negative, training is restarted with random weights.

\algnewcommand\algorithmicforeach{\textbf{for each}}
\algdef{S}[FOR]{ForEach}[1]{\algorithmicforeach\ #1\ \algorithmicdo}

\begin{figure}
 \centering
 \begin{fmpage}{\textwidth} 
  \begin{algorithm}[H]
    \caption{Deep-Opt at a High-Level}

    \begin{algorithmic}[1]
\item[]
  \State  Create set, S, with candidate solutions from uniform distribution.
  \State  Evaluate all solutions in S.
\item[]
  \While {termination condition is not met}
     \State Scale all real evaluations of $s$ to range [0.0,1.0], $\forall(s | s \in S) $
     \State Train a Deep Neural Network (DNN) to map  $s \rightarrow     Evaluation (s), \forall(s | s \in S) $
     \State Freeze the weights of the deep network.
   \item[]
     \State \Comment{\colorbox{yellow} {Stochastically Generate New Candidates}}
     \State Initialize $C$ to be empty.
     
     \Repeat
     \State Generate a new candidate solution, $c$, by randomly perturbing the best solution found so far
     \State Append $c$ to $C$.~~~ $C \gets C + c$
     \Until {(|C| = \emph{Number-To-Generate-From-Model})}

\item[]     
     \State \Comment{\colorbox{yellow} {Back-Drive the Network to Improve Some of the Generated Samples}}

     \State Select a fraction, $F$, of the members of $C \rightarrow C',     s.t. |C'| \le |C|$.
     \State Clamp the network's output to 1.0
     \ForEach {$d \in C'$}
     \State Initialize the inputs of the DNN with $d$
     \State Back-Drive the network until the inputs stop changing $\rightarrow d'$
     \State Replace $d$ in C with $d'$
     \EndFor
\item[]          
     \State Evaluate all candidates, $c \in C$.  
\item[]
     \State \Comment{\colorbox{yellow} {Run the local optimization}}
     \State Select a single candidate, $m$ ($m \in C$), where $m$ is
     has the highest evaluation of all in $C$.

     \State Initialize NASH with $m$.
     \State Run NASH and record all of the unique solutions (into set $Y$)
     evaluated by the NASH run.
\item[]                    
     \State \Comment{\colorbox{yellow} {Update the data to be modeled}}
     \State $Y' \gets$ Select a subset of the best  solutions from $Y$.
     \State $S \gets S + Y'$
     \State Discard Duplicates in $S$; prune $S$ if necessary
\item[]                         
  \EndWhile
  \end{algorithmic}
  \end{algorithm}
  \end{fmpage}

  \caption{Deep-opt: Creating a model of the search landscape to
    initialize fast search algorithms (NASH). We set the
    \emph{Number-to-Generate-From-Model = 50} for all of our tests.
    $|S|$ was kept at 10,000 samples with the size of $Y'$ set at
    1000. For simplicity, $F$ was kept static at 50\% for all of the
    tests in this paper.  }
  \label{fig:deepopt} 
   \vspace {0.0in}  
\end{figure}

There are two implementation details not specified in
Figure~\ref{fig:deepopt}.  In the first step, Line 1, several variants
of creating the initial set, $S$, are possible. The simplest is, as
shown, generating samples entirely randomly.  Fully random sampling gives the broadest
exploration of the search space.  Alternatively, we could have performed a single
NASH run and saved all the points explored.  Using only one run, however, gives a
poor representation of the global landscape because only a deep, single,
path is explored.  A third alternative is to sample a number of seed-points
randomly and also explore their local neighborhoods by making small
perturbations of the seed points.  This gives a cursory indication of
the local landscape around each of the seed points; empirically, this improved results on the problems tested.  We use this initialization method throughout
the paper.  A full analysis was not conducted to measure the
effects of alternate approaches.

Second, as can be seen, there are a large number of parameters and
decisions made in the algorithm design.  Our goal is to show that
using a deep neural network is capable of modeling the search space,
not necessarily advocating a particular network or set of parameters.
Nonetheless, to make the study complete, we need to specify the
networks we used.  For the trials in this paper, two networks were
used.  The first, \emph {Deep-Opt-5}, used a 5-fully-connected-layer
network with 100 hidden units per layer.  The second,
\emph{Deep-Opt-10} used a 10-layer fully-connected network with 20
hidden units per layer and skip connections between every layer and
its predecessors.  Both networks have a single output - the estimate
of the evaluation of the function being estimated.  Weight decay was
used in training (0.98).  A variety of networks were tried, ranging
from simple single layer networks to even deeper ones.  As expected,
the best performer was dependent on the problem, however
these provided good results across multiple problems.


Finally, the very careful reader may have noted that the effort to
separate the steps in Lines 8-11 from those in lines 13-19. Later in
the paper, we will show how lines 8-11 can be replaced with other
procedures.

\subsection {Visualizing the Learning}

Before turning our attention to the empirical tests in the next
section, we present a motivating example to demonstrate how the local
search algorithms utilize the models created.  We examine a simple
problem in which the evaluation is: $ evaluation (\boldsymbol{x}) =
\sum_{i=1}^{50} ( x_i * sin(x_i)) $.  We instantiate this problem
with 50 parameters that can take on values between [0,100].  Note that
in this maximization problem, each parameter is independent.  Though
this is trivial problem, it serves to demonstrate the expected
behavior.

If a NASH search is initialized with a solution vector that has low
values for any of the parameters, there are many local optima on the
way to the global optima.  Nonetheless, because of the ease of the
problem, even with no learning, some of the runs will find the global
optima.

\begin{figure}[t]
\centering
\includegraphics[width=0.375\textwidth]{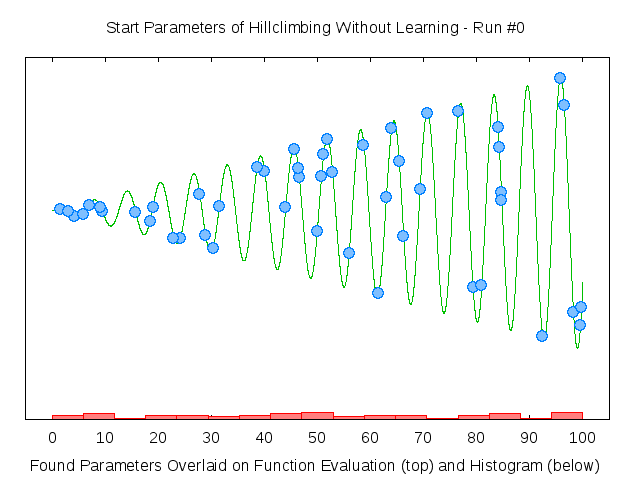}
\includegraphics[width=0.375\textwidth]{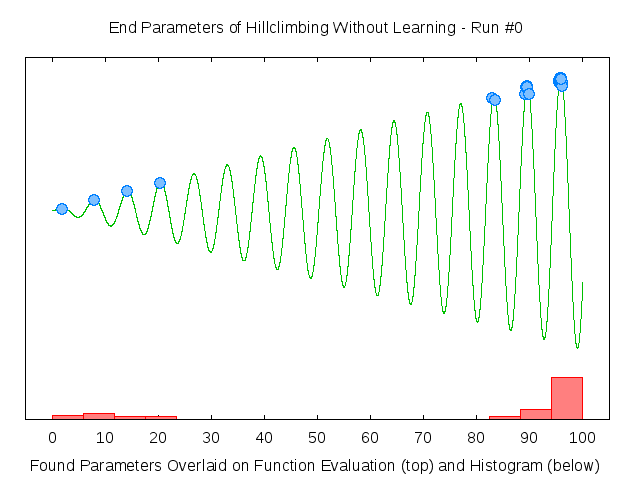}
\\
\includegraphics[width=0.375\textwidth]{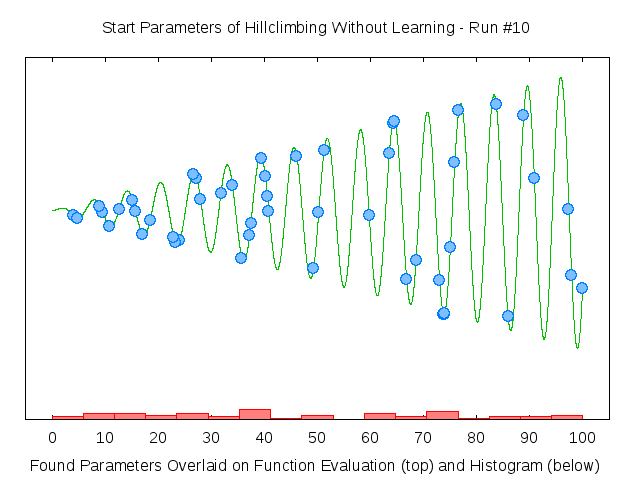}
\includegraphics[width=0.375\textwidth]{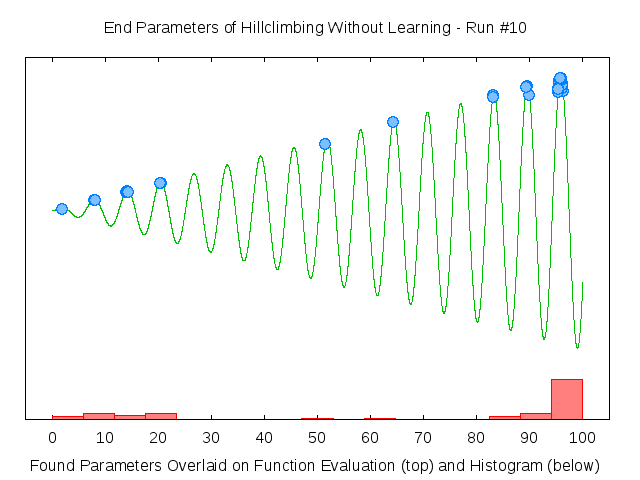}
\\
\includegraphics[width=0.375\textwidth]{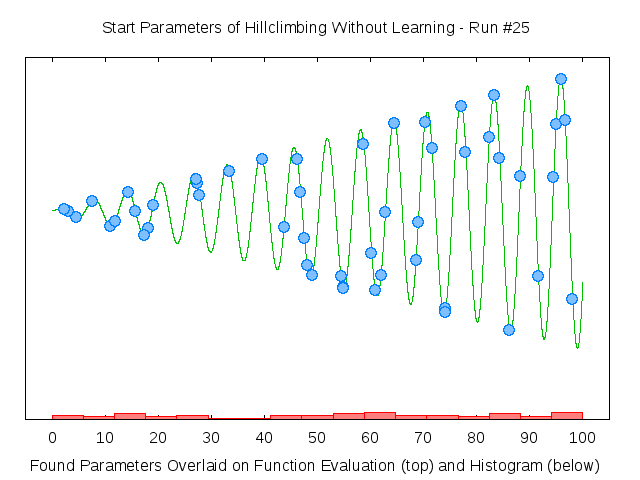}
\includegraphics[width=0.375\textwidth]{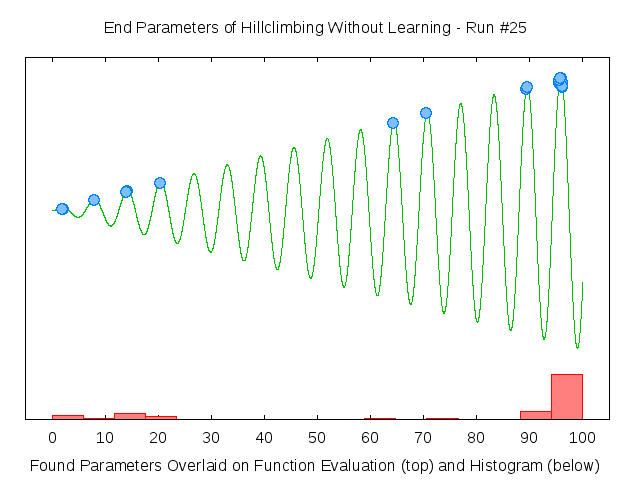}
\\

\caption{Sines Problem \underline{without Learning}. The (green) line is the
  underlying function to be maximized.  The (blue) points are the
  settings of each the 50 parameters in the beginning (left) and
  ending (right) of the NASH procedure (random perturbations of best
  seen-so-far).  A histogram showing the distribution of the points is
  shown in red (this is shown because of the difficulty in seeing the
  overlapping points in this graph).  Note that in the beginning of
  each run (left column), the points are uniformly distributed across
  the full input space. The results have vastly improved through
  hillclimbing, though not all points have made it to the global
  optimum. Top Row: NASH run \#1, Middle Row: NASH run \#10 (this is
  the HC run made after the 9th modeling step), Bottom Row: NASH run
  \#25. Note about the graph: although each points represents one of
  the 50 parameters in the same single solution string, they are shown
  'folded-over' onto the same graph.  This is possible because each
  parameter is independent and is evaluated with respect to the same
  function.}

\vspace {0.5in}
\label{fig:sineHc}
\end{figure}

In Figure~\ref{fig:sineHc} (top-row), in the left column are the
starting points for all 50 of the parameters before the NASH algorithm
is run. As expected, they are randomly distributed across the input
range.  By the end of the NASH run (right column) many of the
parameters are close to their optimal settings.

The middle row of Figure~\ref{fig:sineHc} shows the same (start/end
configurations) for the 10th restart of NASH.  Because NASH is still
initialized randomly (no learning), we expect to see largely the same
distribution of points in the beginning and the end.  The same is
shown in the bottom row, the 25th restart of NASH.

\begin{figure}
\centering
\includegraphics[width=0.375\textwidth]{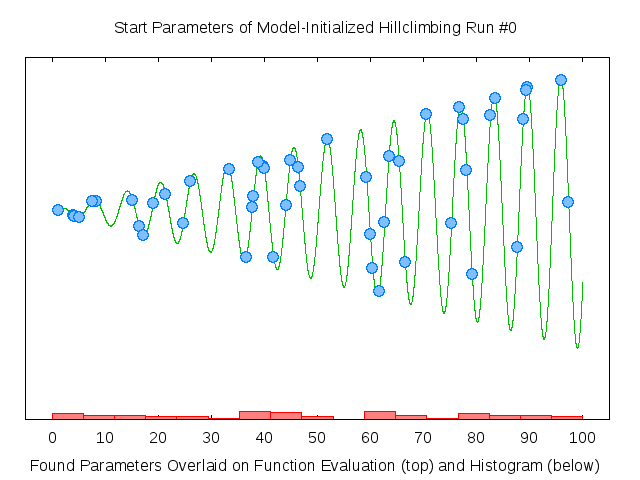}
\includegraphics[width=0.375\textwidth]{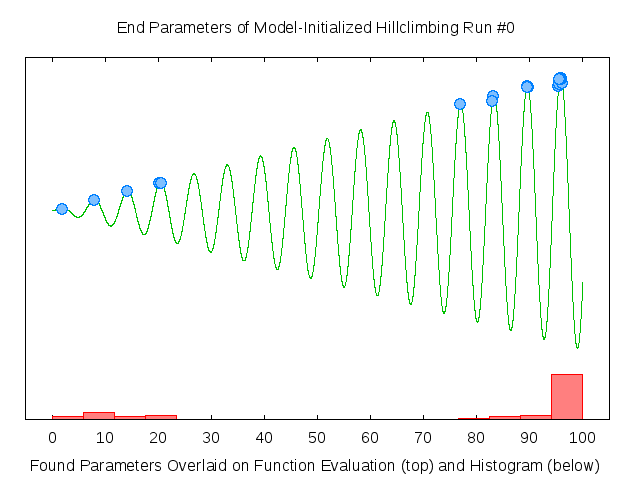}
\\
\includegraphics[width=0.375\textwidth]{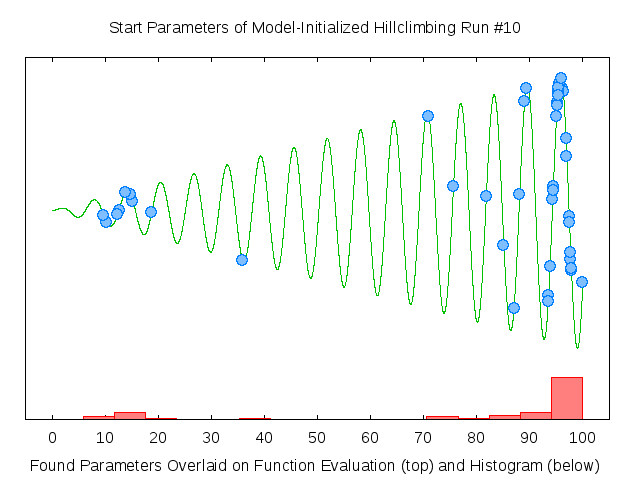}
\includegraphics[width=0.375\textwidth]{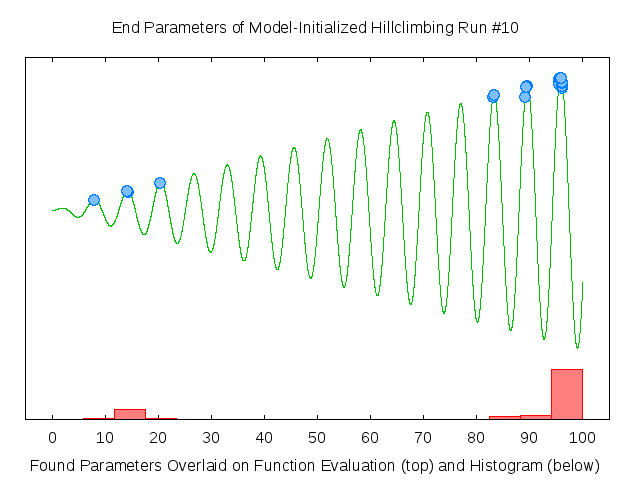}
\\
\includegraphics[width=0.375\textwidth]{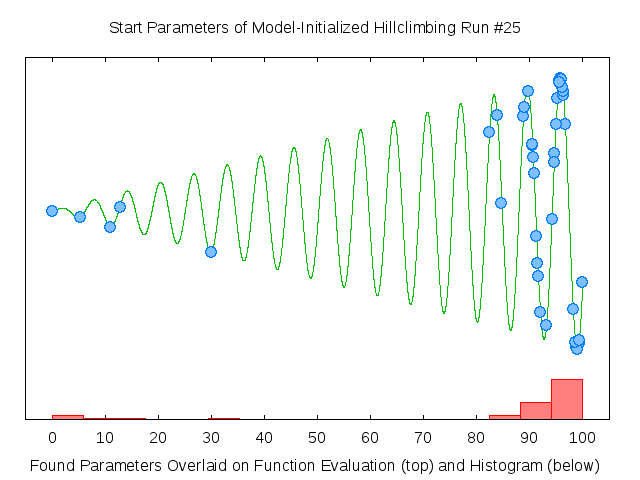}
\includegraphics[width=0.375\textwidth]{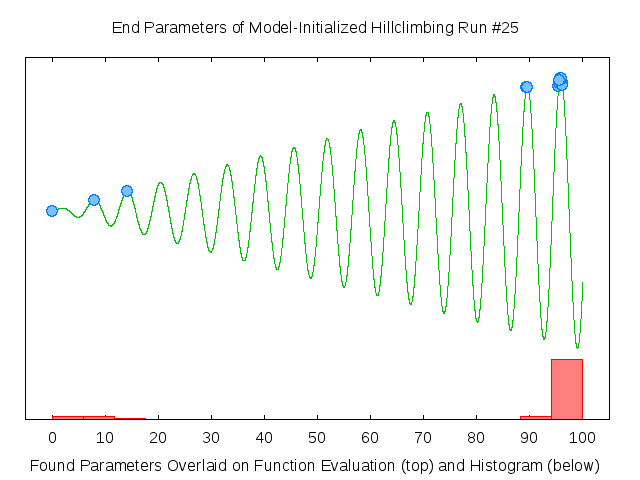}
\\

\caption{Sines Problem with Search-Space Modeling / Learning.  Note
  that even at the beginning of the runs, shown in the left column of
  rows 2 \& 3 (run \#10 \& \#25 respectively), the hillclimbing
  \emph{begins} in regions of high performance -- thereby leading to
  better performance overall by the end of the hillclimbing search.
  In run 1 (top row, left)- since no learning has yet happened, the
  points are randomly distributed in the beginning of the run.  }

\vspace {0.5in}
\label{fig:sineModel}
\end{figure}

Next, we repeat the same experiment with Deep-Opt.  The only
difference is in the initialization of the NASH algorithm. The samples
that are generated in each NASH run are added to the set of solutions
that are modeled by the neural network. From this neural network
model, $M$ new samples are drawn. The single best of the $M$ is used
to initialize the next NASH run.  As expected, in
Figure~\ref{fig:sineModel}, the first run looks similar to the earlier
case with no model.  This is because there is no information in the
model as yet.  However, after the 10th NASH restart, (middle-row,
Figure~\ref{fig:sineModel}) there is a distinct difference in the
starting values -- many more are already in high-evaluation regions.
The best solution found through sampling had many of its parameters in
the right region of the search space.  The likelihood of these
reaching the global optimum is increased through NASH (right column).
By the 25th run, this trend is even further evident.  Sampling the
model works as expected: the starting point for the hillclimbing is
already in a better region - where more parameters are closer to the
global optima.

Though this problem is simple, it demonstrates how the modeling can
improve the search results.  Interestingly, early on in our studies,
modeling \emph {sometimes led to poorer} overall performance.  Why?
As the probabilistic model improved, exploration decreased -- more
samples started in a basin of attraction that led to the same local
maxima as was seen previously.  Improvement slowed because the updates
to the model all happened with similar candidate solutions.  The
algorithm parameters, in particular the size of the sample set that
was used for modeling, $|S|$, were tuned to the settings shown in
Figure~\ref{fig:deepopt} to slow the convergence; this vastly
improved performance.  The effects of output scaling (discussed later
in Section~\ref{scaling}) are also relevant to this observation.

\clearpage
\section{Empirical Results}
\label{tests}

In this section, we examine the benefits of the probabilistic model to
select starting points for the NASH optimization procedure on a number
of problems drawn from the literature and real-world needs. As
described earlier, when the model is used, $M$ samples are generated,
the best of which is used to initialize NASH.  To ensure that the
model is actually providing useful information and that it is not
merely the process of examining $M$ samples before beginning a new run
of NASH that is yielding the improved performance, three variants of
NASH are explored.  Though they vary in seemingly small implementation
details, the effects on performance can be dramatic.

\begin{itemize}

\item NASH-V1: This is exactly NASH shown in Figure~\ref{fig:nash}.

\item NASH-V2: Before beginning the NASH run, $M$ samples are
\emph{randomly} generated and evaluated.  The best one found from the
$M$ generated is used to initialize the NASH algorithm.  No learning
is used here.  This variant is included to test whether just the
process of generating multiple samples and selecting the best prior to
starting NASH is enough to provide improvement to the final result --
even with no modeling.

\item NASH-V3: Before beginning the NASH run, $M$ samples are
  generated by making small perturbations to the best solution found
  in all the previous NASH runs.  (The first NASH run is initialized
  randomly).  Each of the $M$ samples is then evaluated.  The best
  sample found from the $M$ is used to initialize the NASH
  algorithm. This variant tests whether the neural network models
  actually capture the shape of the search space, or whether they are
  only (inefficiently) forcing search around the best solutions seen
  so far.

\end{itemize}

It is interesting to note that on some of the problems detailed below,
adding any of these heuristics will lead to a \emph{degradation} in
performance.  For these problems, it is likely that the search space
does not contain easily learnable trends-- either because large
portions are flat or pocked with local-optima, or that the information
cannot be correctly modeled with the networks used here.  This will be
discussed with the problem descriptions.  Using problem-specific
hand-crafted features, such as done with
STAGE~\cite{boyan2000learning}, may help.


The termination condition for each NASH run was either (1) 10,000
evaluations were performed or (2) 500 evaluations were conducted with
no-improvement.  The latter indicated that the search might be trapped
in a local maxima.  All approaches were given a total of 500,000
evaluations.  The number of NASH runs that were conducted within the
500,000 evaluations was dependent on the problem and how quickly/often
the algorithm was unable to escape local optima.

\subsection{Noisy Evaluations}

In the previous section, we used a simple Sines maximization problem
as an illustrative example of how learning aids search.  Due to the
problem's simplicity, most search algorithms can perform well on it.
However, with the introduction of noise, a clearer separation in
performance emerges.

In this version of the problem, \emph{Noisy-Sines}, the evaluation was
modified to include significant uniformly distributed random noise.
Uniformly chosen random noise between $[0,0.5]$ was added to the
evaluation.  For large parts of the search space, the overall
evaluation is dominated by noise (note that if the same solution is
evaluated twice, the noise is chosen independently each time).  

\begin{equation}
  evaluation (\boldsymbol{x}) =  \dfrac{\sum_{i=1}^ {50} ( x_i * sin(x_i))} {50.0 * 100.0}    + uniformNoise (0,0.5)
\end{equation}

The performance of each algorithm is judged by the best solution found
for the underlying objective function (as determined \textbf{without
  the noise}); no algorithm is privy to the underlying real function.
The results are shown in Table~\ref{noisySines}.  For each of the 5
algorithms tried (NASH-1,NASH-2,NASH-3, Deep-Opt-5-Layers,
Deep-Opt-10-Layers), the best evaluation averaged over trials is
listed in the first row. The last two rows show the significance of
the difference between the algorithm's performance and the performance
of Deep-Opt-5-Layers and Deep-Opt-10-layers, respectively.

\begin{table}[h]
  \centering
    \vspace {0.1in}
    \caption{Results for Noisy Sines Problem (20 Trials for Each Approach)}
    \label{table:seating}
    \begin{tabular}{llll||ll}

      \toprule
    &NASH-1& NASH-2& NASH-3& Deep-Opt-5 & Deep-Opt-10\\
\midrule
   Overall Average & 0.690 & 0.691 & 0.700 & 0.731 & 0.726 \\
   Difference to Opt-5 Signif? & >99\% & >99\% & >99\% & - & 96\% \\
   Difference to Opt-10 Signif?& >99\% & >99\% & >99\% & 96\% & - \\

\bottomrule
    \end{tabular}
    \label{params}
    \vspace {0.05in}
    \label{noisySines}      
\end{table}

\subsection{Stable Marriage \emph{Reception-Party Seating}}
\label{seating}

Though this problem shares part of its name with the stable marriage
problem, it is more akin to knapsack/packing problems.  Real versions
of this problem have arisen in topics as diverse as processor
scheduling to intern and group seating layouts.

In a canonical version of this problem, $G$ parties are invited to a
formal-seated party, such as a wedding reception.  Each party can have
a variable size.  Each member of the party must sit together at one of
the $T$ tables, which each have a capacity $C_t$.  The additional twist to
this problem is that each $G$ has a preference with whom to sit with,
expressed as a real value.  The full preference matrix is $| G \times
G | $.  Preferences can be negative, and not constrained in magnitude.
Further, it is not a requirement that each guest express a preference
to every other guest, or even to any guests.  Preferences may not be
symmetric.

The goal is to find a seating assignment that (1) keeps the members of
each group together, (2) does not seat people beyond the capacity of
the table, (3) maximizes the summed happiness/preferences over all the
tables. For the size of the problems explored here, the reception has
10 tables, each with capacity 12 people.  Each guest's party is
randomly chosen between 1 and 3 people.  Preferences were expressed as
a value between [-100, +100].  The number of groups, $|G|$, was set to
50.

To encode the solution as a vector, each group was assigned T
parameters, corresponding to each of T tables; there were a total of
500 $(50 \times 10)$ parameters, ($realValueParameter_{g,t}$).  At
evaluation time, these 500 parameters were sorted from high to
low. Based on the sorted list of $realValueParameter_{g,t}$
assignments were made in order from highest to smallest of group $g$
to table $t$.  Note that the assignment occurred only if the group was
(1) as yet unseated and (2) the table could hold the size of the
group; otherwise that parameter was ignored and the next one
processed.  This encoding has the benefit of not only specifying 
each groups' preferences to tables, but also being able to encode
``how important'' it is that a particular group be assigned to a
particular table.

20 unique problems were created and tested with randomly generated,
complete, $|G \times G|$ preference matrices.  The random generation
of problems led to an extremely large spread of final answers across
problems.  To summarize the results, we compared the five approaches,
and gives the numbers of problems (out of 20) on which each algorithm
obtained the highest evaluation (highest summed preferences at the
tables, with all the constraints being met).  The results are shown in
Table~\ref{stableMarriage}. The next line of the table give the number
of trials (out of 20) that Deep-Opt-5-layers outperformed the other 4
methods.  The last line does the same for Deep-Opt-10-layers.

\vspace {0.1in}
\begin{table}[h]
    \centering
    \caption{Stable Marriage Reception-Party Seating:  Number of times
      each algorithm outperformed all others (out of 20)}
    \label{table:seating}
    \begin{tabular}{llll||ll}    
    \toprule
    &NASH-1& NASH-2& NASH-3& Deep-Opt-5 & Deep-Opt-10\\
    \midrule
Overall Best (Out of 20) & 3 & 2 & 0 & 10 & 5 \\
Deep-Opt-5 Outperformed(ties) & 14 & 16 & 19 & - & 11 (1) \\
Deep-Opt-10 Outperformed(ties) & 10 & 13 & 15 & 8 (1) & - \\
    \bottomrule
    \end{tabular}
\label{stableMarriage}
\end{table}

\vspace {0.3in}

~

\subsection{Graph Bandwidth}   

Given a graph with $V$ vertices and $E$ edges, the graph bandwidth
problem is to label the vertices of the graph with unique integers so
that the difference between the labels between any two connected
vertices is minimized.  Formally, as described
in~\cite{wiki_bandwidth}, label the $p$ vertices $v_i$ of a graph $G$
with distinct integers $f(v_i)$ so that the quantity
$\max\{\,|f(v_{i})-f(v_{j})|:v_{i}v_{j}\in E\,\}$ is minimized ($E$ is
the edge set of $G$).  More details of the complexity of this problem
can be found in~\cite{chinn1982bandwidth}.  Interest in this problem
stems from a variety of sources, including constraint
satisfaction~\cite{zabih1990some} and minimizing propagation delay in
the layout of electronic cells.

The solution is encoded as follows: each vertex is assigned a
real-valued parameter (full solution encoding of length $|V|$).  The
vertices are sorted according to their respective assigned values.
The integers $[1 .. V]$ are then assigned to the vertices in their
respective sort position. Once each vertex has an integer assignment,
the maximum difference between the assignments of connected edges is
returned.  The results are shown in Table~\ref{graphBandwidthResults}.
This is a particularly difficult problem; ties are shown in
parentheses.

\begin{table}[h]
    \centering
    \caption{Graph Bandwidth:Number of times
      each algorithm outperformed all others (out of 20)}
    \label{table:seating}
    \begin{tabular}{llll||ll}    
      \toprule

    &NASH-1& NASH-2& NASH-3& Deep-Opt-5 & Deep-Opt-10\\
      \midrule
      
      Overall Best & 0 & 0 & 6 & 10 & 8 \\
      & & & & \\                  
      Deep-Opt-5 Outperformed & 20 & 20 & 12 (3) & - & 6 (6) \\
      Deep-Opt-5 Difference Signif?& >99\% & >99\% &96\% & - & no \\      
      & & & & \\                  
      Deep-Opt-10 Outperformed & 20 & 20 & 12 (2) & 8 (6) & - \\
      Deep-Opt-10 Difference Signif?& >99\% & >99\% &96\%& no & - \\
    \bottomrule
    \end{tabular}
  \label{graphBandwidthResults}
\vspace{0.2in}
\end{table}

\subsection{Graph-Based Constraint Satisfaction}
\label{realconstraints}

Constraint Satisfaction has numerous real-world applications.  We
recently used it for resource allocation and job scheduling.  Is is
presented here in its simplest form.

Simply stated, in this problem, there are $P=100$ real-value
parameters in the range [0,1.0].  These parameters are assigned to the
vertices in a graph. The graph contains 2,000 randomly chosen,
directed, edges which specify a constraint that the origination-node
must hold a value greater than the destination-node.  The optimization
problem is to assign values to the nodes such that as many of the
2,000 constraints are satisfied as possible.  If the constraint is not
met, the error is the absolute difference in the two values.  The
error, to be minimized, is summed over all constraints.  The results
are shown in Table~\ref{realConstraintResults}.

\begin{table}[h]
    \centering
    \caption{Real Valued Graph-Based Constraint Satisfaction: Number of times
      each algorithm outperformed all others (out of 20)}
    \label{table:seating}
    \begin{tabular}{llll||ll}    
      \toprule

    &NASH-1& NASH-2& NASH-3& Deep-Opt-5 & Deep-Opt-10\\
      \midrule
      
      Overall Best & 0 & 0 & 2 & 13 & 5  \\
      & & & & \\            
Deep-Opt-5 Outperformed & 20 & 20 & 16 & - & 14 \\
Deep-Opt-5 Difference Signif?& >99\% & >99\% & >99\% & - & no  \\
      & & & & \\            
Deep-Opt-10 Outperformed & 20 & 20 & 17 & 6 & - \\
Deep-Opt-10 Difference Signif?& >99\% & >99\% & >99\% & no  & - \\

    \bottomrule
    \end{tabular}
    \label{realConstraintResults}
\end{table}

\vspace{0.4in}
~

\subsection {Graph-Based Discrete Constraint Satisfaction}

In this variant of the previous graph-based constraint satisfaction
problem, the exact same setup is used as in
Section~\ref{realconstraints}, however, each node may only take on 1
of 16 letters -- ${A..P}$.  In terms of the real-world application of
job scheduling mentioned above, in this version of the problem, jobs
can enter the system only at specific, synchronized times.  This makes
the problem closer to a selection problem (where one of the 16 values
is selected for each of the nodes) as compared to the previous
instantiation where a real value was assigned to each node.

Though conceptually a small difference from the above encoding,
discretization has enormous ramifications in the solution encoding.
The simplest encoding is to use 100 real-valued outputs, one for each
node, and divide the [0,1] region into 16 evenly spaced regions, each
assigned to a single letter.  However, for the reasons explained in
Section~\ref{discrete}, this encoding performs poorly.  Instead, we
use an encoding more amenable to selection problems and/or
discrete-parameters; this encoding improves the performance of both
Deep-Opt and as well as NASH alone.

The encoding used is similar to the Reception-Party-Seating task
described in Section~\ref{seating}.  Each vertex in the graph is
assigned 16 real-valued parameters; each corresponding to a single
letter ${A..P}$ (In contrast, recall that with the encoding described
in Section~\ref{realconstraints}, each vertex was simply assigned 1
real-valued parameter).  In each set of 16, the maximum value is found
and the corresponding letter assigned to the vertex.  In sum, for a
100 node graph, 1,600 parameters are used.  Once the graph nodes are
assigned values, the rest of the evaluation proceeds as described in
Section~\ref{realconstraints}.

\vspace{0.2in}    

\begin{table}[h]
    \centering
    \caption{Discrete Valued Graph-Based Constraint Satisfaction:
      Number of times each algorithm outperformed all others (out of
      20)}
    \label{table:seating}
    \begin{tabular}{llll||ll}    
      \toprule
    &NASH-1& NASH-2& NASH-3& Deep-Opt-5 & Deep-Opt-10\\
      \midrule
      Overall Best & 0 & 0 & 3 & 13 & 4 \\
      & & & & \\                        
     Deep-Opt-5 Outperformed & 19 & 19 & 16 & - & 14 \\
     Deep-Opt-5 Difference Signif?& >99 \% & >99 \% & 98\% & - & no  \\
      & & & & \\                        
     Deep-Opt-10 Outperformed & 19 & 19 & 14 & 6 & - \\
     Deep-Opt-10 Difference Signif?& >99 \% & >99 \% & 92\% & no & - \\
     
    \bottomrule
    \end{tabular}
    \label{params}
\vspace{0.2in}    
    
\end{table}

\subsection{Two Dimensional Layout Problems}
\label {2d}

This section highlights the limitations of the Deep-Opt approach.  A
number of problems which broadly encompassed the task of two
dimensional layout, did not improve significantly with search space
modeling.  Two of the problems are detailed here.

\subsubsection {Minimizing Crossings}

The goal is to find a planar layout of a graph's nodes that minimizes
edge crossings. See~\cite{wiki_crossing} for more details.  In
general, the edges can be drawn in any shape.  For simplicity, in our
implementation, the edges are drawn only with straight lines, this is
termed the \emph{rectilinear crossing number}.

For our tests, each node was represented with two parameters (x,y
coordinates).  Small graphs were tried with 25 nodes.  This yielded a
solution encoding of 50 real-values, which specified the coordinates
of each point on a plane.  Each graph had 50 randomly chosen
connections.  20 randomly generated problem instantiations were
attempted.

One of the interesting findings is that NASH-2 outperformed NASH-3. In
most previous experiments, this was reversed.  NASH-2 received a
higher score in 12 out of the 20 problems (the scores, as measured by
a standard t-test were statistically different with $p=0.96$).  Recall
that NASH-2 initialized each hillclimbing run by first generating a
small number of random candidate solutions and selecting the best one.
In contrast, NASH-3 perturbs the current best solution to determine
the best starting point for NASH.  Although left for future
exploration, it is worth investigating what insight this gives about
the search space?  If searching around the current best does not yield
as good results as randomly starting over, does the search space have
more or less optima, or are the local optima further spread apart,
deep, etc?  We leave the speculation of the ramifications of NASH-2
outperforming NASH-3 to future work.  However, because the Deep-Opt
can just as easily be applied to NASH-2 as NASH-3, for the experiments
in this section, we used it to ``wrap'' NASH-2.  Everything else, all of the
parameters, etc., remained the same.

\begin{table}[h]
    \centering
    \caption{Results for Minimizing Crossing on 20 randomly generated problems}
    \label{table:mincross}
    \begin{tabular}{llll||ll}    
      \toprule

    &NASH-1& NASH-2& NASH-3& Deep-Opt-5 & Deep-Opt-10\\
      \midrule
      
      Overall Best & 1 & 5 & 3 & 8 & 3 \\
      & & & & \\      
      Deep-Opt-5 Outperformed & 16  & 11 & 14  & - & 15 \\
      Deep-Opt-5 Difference Signif?& >99\% & no &97\% & - & no \\      
      & & & & \\
      Deep-Opt-10 Outperformed & 14 & 6 & 13 & 5 & - \\
      Deep-Opt-10 Difference Signif?& >97\% & no & no & no & - \\

    \bottomrule
    \end{tabular}
  \label{params}

\end{table}
\vspace {0.1in}

\subsubsection {Image Approximation via Triangle Covering}
\label{triangles}

This is the only problem in the paper for which the parameters for
NASH were changed to be optimized for this problem.  Accordingly,
Deep-Opt also used the same parameters.  Had these parameters not been
reset, neither NASH nor Deep-Opt would perform as well as shown here,
with Deep-Opt \emph{under-performing} NASH alone.

In this problem, there is an intensity target image (just black and
white, no color information), $I$, that is $N \times N$ pixels.  The
goals is to find T triangles and their intensities that approximate
the image.  Specifically, each triangle must specify three vertices
between [0,N] in both X\&Y and an intensity value.  The triangles are
drawn onto an initially empty canvas.  The triangles may overlap;
their intensities are additive.  After all T triangles are drawn, the
resulting image is scaled back into the appropriate space (0 .. 255
pixel intensities), and compared, pixel-wise, to the original image.
The $L_2$ distance is to be minimized.  (The problem is set up as a
maximization problem by taking the the reciprocal of the summed $L_2$
distance.)  This is a particularly difficult/interesting problem when
T is small.

\begin{figure}[t]
\centering
\includegraphics[width=0.35\textwidth]{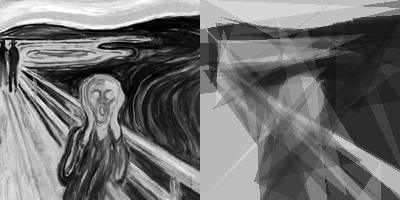}
~\\
~\\
\includegraphics[width=0.35\textwidth]{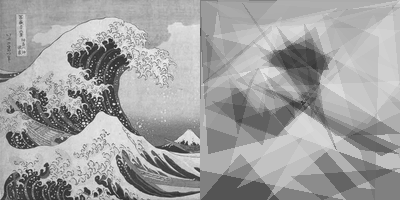}
~\\
~\\
\includegraphics[width=0.35\textwidth]{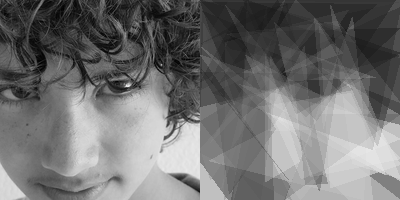}
~\\
~\\
\includegraphics[width=0.35\textwidth]{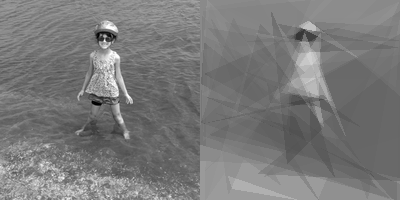}

\caption{In each of the 4 sets of images, the left of the pair is the
  original image.  On the right is the reconstruction of the image using
  the 50 triangles found by Deep-Opt.}

\label{fig:triangles}
\end{figure}

To set the parameters for NASH and Deep-Opt, we tested this problem
with 50 triangles, trying to approximate an intensity based crop of
``The Scream'' by Edvard Munch.  Each triangle was encoded as 7
parameters: 6 for the (x,y) coordinates of three vertices and 1 for
the intensity.  With 50 triangles, there were a total of
($50\times7=$) 350 parameters in the solution encoding. Once the
parameters were set, 3 other images, shown in
Figure~\ref{fig:triangles}, were also attempted with the same
parameter settings.  The relative performance is given in
Table~\ref{table:triangle}.  Note that the overall performance is
virtually identical.  This is averaged over 10 trials on each image.

\vspace{0.2in}
\begin{table}[h]
    \caption{NASH vs. Deep-Opt on Triangle Covering - Equivalent
      Performance over Multiple Runs \& Images}
      \label{table:triangle}
    \centering      
    \begin{tabular}{p{4cm}p{4cm}p{4cm}}
    \toprule
    Image & Performance of NASH & Performance of Deep-Opt\\
    \midrule
    The Scream & 1.00 & 1.00\\
    Great Wave off Kanagawa & 1.00 & 1.001 \\
    Photograph 1 & 1.01 & 1.00 \\
    Photograph 2 & 1.00 & 1.001 \\
    \bottomrule
    \end{tabular}

\end{table}
\vspace{0.2in}

\clearpage

\section{Extensions \& Alternatives}
\label{alternatives}

In this section, we briefly describe three extensions and further
tests to the Deep-Opt.  Though the results are promising, they are
preliminary.

\subsection {Discrete/Binary Parameters}
\label{discrete}

Deep-Opt has been described with real-valued parameters.  The input
parameters, as well as the target output, are scaled to [0.0,1.0].  However,
there are a wide variety of problems that employ discrete or binary
parameters.  Here, we present one method to tackle such problems.

To ground the discussion, let's examine a simple \emph {two
  dimensional checkerboard problem}~\cite{baluja1998fast}.  In this
problem, there is a planar $15\times15$ grid of binary digits.  The
goal is to set each digit such that it is the opposite of the digits
in its 4 primary directions.  There are two globally optimal
solutions, but there there are also many locally optimal solutions
such that any single bit-flip will not yield improved performance.

First, we conducted an experiment to determine how Deep-Opt, as
described to this point, would perform on this task.  The solution
encoding is 225 bits.  In this instantiation, the maximum possible
evaluation for this task is 676 (13*13*4); each of the bits in the 4
primary directions that are correctly set for the inner $13\times13$
square contributes one point to the evaluation.  The average Deep-Opt
based solution quality was 537.  For comparison, the three versions of
NASH, without any learning, had average solution qualities of 655,655
and 668 (very close to the optimal), respectively.  This difference
between these and Deep-Opt is one of the largest witnessed in this
entire study.

Why did this happen?  In this initial attempt, the straight-forward
method to interpret the real valued solution parameters as binary was
used.  If the real value was over 0.5, the bit was assigned to 1 and 0
otherwise.  With this scheme, note that small changes to a parameter's
real value often \emph{did not yield any change} in solution string
(unless the value was close to 0.5) or, thereby, to its evaluation.
Therefore, many solution strings appeared to have the same evaluation,
though they held different values as real-valued vectors.

Ideally, if the bit position should be 1, we should favor solutions
that have the real value associated with that bit position as far
above 0.5 as possible (and the opposite for bit positions that should
be 0).  To do this, we use a technique similar to stochastic sigmoid
units in training the neural network to model the search space (for a
recent paper on this and related topics, see
~\cite{raiko2014techniques}).

Recall that in the candidate generation phase (Lines 13-19 in
Figure~\ref{fig:deepopt}), we start with an input of a candidate
solution vector, $s$.  Over a number of iterations of back-driving the
network, the inputs are modified by following the gradients needed to
transform the candidate solution to one such that the network computes
a value of 1.0 when the revised candidate solution is input.  In the
discrete version (referred to as Deep-Discrete-Opt), we instead treat
each real valued parameter, $p, p \in s$ as a probability.  We
generate a small number of binary solutions strings (200) where each
parameter in position-$p$ has a probability of being assigned a 1.0 of
$p$.  All of these binary solutions are then fed to the network and
the gradients computed for all of the solution strings (to move the
network's evaluation of each of them closer to the target output of
1.0).  Note that there is no need to evaluate each of these against
the actual, real, evaluation function -- only the network's output is
measured to set the error signal.  Although this procedure adds to the
processing time, the result are vastly improved.  Out of 10 trials, 9
had perfect evaluations of 676; overall the trials had an average of
673.4.  This overcomes the previous limitations of the naive
implementation of binarizing the real-value parameters.

For completeness, we also tried an alternative to this procedure to
determine if just increasing the learning rate in the generation process
would cause enough moves across the 0.5 boundary to achieve the same
benefits.  A variety of larger learning rates were tried (from $2\times$ to
$40\times$ the standard learning rate used previously).  In no case did the
performance improve.  The generation of binary strings outperformed all
versions with only increased learning rates.

\subsection {The Role of Scaling Examples and Their Outputs}
\label{scaling}

Deep-Opt, as used in this paper, works by back-driving the neural
network inputs to change to cause the network to yield an output of
1.0 at its output node.  Recall that the evaluations of the solutions
generated in $S$ are scaled, before each training session, to values
between [0.0,1.0].

Scaling the evaluations of $S$ to a fixed range leads to a subtle
complexity.  Note that a subset of the members of $S$ change in every
training cycle: as NASH completes a hillclimbing run, some of the
newly found solutions are appended to $S$ and other solutions from $S$
removed.  Often, in the early parts of search, the range of the actual
evaluations present in $S$ expands to include new, higher evaluation
members while many of the randomly found initial members of $S$ (with
low evaluations) are still present.  The opposite happens in the
latter parts of search; the range of values contracts as the search
focuses primarily on the better solutions and all the solutions have
close (high) evaluations.

Because of this, note that in trainingCycle$_t$, a member, $s$ may
have a real evaluation of X that is scaled to X' (when all the members
of $S$ are scaled between 0 and 1.0).  In trainingCycle$_{t+1}$, that
same $s$, which has the same real evaluation of X, may be scaled to
X'', where X'' could be either greater or less than X'.  Because the
networks continue their training in every cycle, this has not proven
to be a problem as the change in evaluation seems to be quickly
recovered; this may be because the relative ordering of the samples
does not change.  However, in the future, it will be interesting to
try variants in which the re-scaling is not done.  This may require
engineering the setup of the problems so that they always output
values in a fixed, \emph{a priori} known, range.

\begin{figure}
\centering
\includegraphics[width=0.7\textwidth]{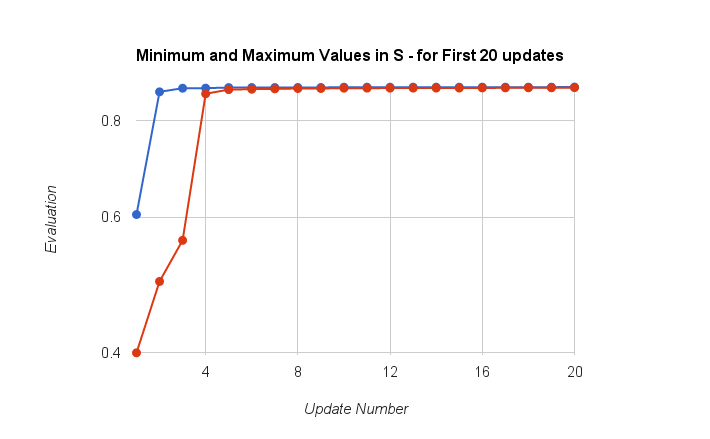}
\caption {Notice the rapidly diminished difference in evaluations
  between the best and worst members of $S$ as search progresses.  $S$
  is the set of candidate solutions from which the probabilistic model
  is created.}
\label{fig:convergence}
\end{figure}

This expansion and contraction dynamic process happens implicitly
through the addition of new members of $S$ as search progresses.
Returning to the Sine problem, we can witness the convergence of the
minimum and maximum values represented in $S$ as new items are added
to $S$ (shown with a small $|S|$ =5000); see
Figure~\ref{fig:convergence}.  The important trend is that the minimum
and maximum values rapidly converge.  This \emph{does not} necessarily
imply that all the candidate solutions in $|S|$ are similar (though
likely many are); it does mean that they are close in
performance~\footnote{ Though difficult to see, towards the right of
  the graph, when new good solutions are found in later updates, the
  difference between the lines increases for a few iterations.}. The
rate of this convergence corresponds to the exploitation
vs. exploration trade-off.

One can imagine that instead of scaling the target outputs to values
in the range [0.0,1.0], they were scaled to [0.0,$Z$], where $Z<1.0$.
In this approach, the highest scoring $s \in S$ will have a value of
$Z$.  When the network is back-driven, it is still driven \textbf{to
  find solutions that produce a 1.0} in the output.  Semantically,
this attempts to create new solutions that are \emph{explicitly better
  than}, not just equal to, those seen so far.  The success of this
approach is pinned on the network's successful extrapolation of the
underlying search surface to regions of \emph{better} performance.
Many versions of this were tried, where $Z$ was set to
[0.2,0.5,0.8,0.9,0.95, and 0.99] in various experiments.  Although the
results are preliminary, setting the $Z$ to 0.95 and higher had little
effect on the results, when compared to setting $Z=1.0$.  Setting $Z$
in the low range often hurt performance.  Further exploration is left
for future work.

\subsection {Alternative Underlying Search Algorithms}

In this paper, we have coupled the use of deep-net modeling with an
extremely simple, fast, localized search algorithm, NASH.  It is
important to also consider the possibility of using alternate search
algorithms with the same modeling procedures.  It is obvious how
alternative search heuristics such as simulated
annealing~\cite{kirkpatrick1983optimization} and TABU
search~\cite{glover1989tabu} can easily be substituted as they are
often used to search neighborhoods around a single point in a manner
similar to NASH. Although we will not delve into the general debate of
the merits of these simple techniques with more sophisticated search
techniques, as this has been an active area of discussion for several
decades
~\cite{juels1996stochastic}~\cite{lang1995hill}~\cite{baluja1995empirical}~\cite{mitchell1994}~\cite{ross1995comparing}~\cite{lobo2015hillclimbers}~\cite{de1993genetic},
it is interesting, to consider the use of the probabilistic models
with very different search paradigms, such as genetic algorithms.

Recall that, unlike NASH, genetic algorithms work from a population of
points.  Members of the population are created with recombination
operators (crossover) that combine the elements of two or more
candidate solutions.  The newly created solutions are further randomly
perturbed (mutated) to reveal the 'children' solutions that are the
candidate solutions to evaluate next~\cite{goldberg1989genetic}.
Numerous variations of GAs and task-specific operators are possible
and have been explored in the research literature.  Next, we perform a
set of tests using a a simple-GA with the parameters shown in
Table~\ref{table:gaParms}.  These are typical of GAs used for static
optimization problems in the literature\footnote{It is likely that
  other operators and operator application rates may yield improved
  optimization algorithms for these problems. Our intent is more
  modest- simply to show that the Deep-Opt framework can be as easily
  wrapped around multiple-point search-based algorithms, such as GAs,
  as well as single-point search based algorithms, such as
  Hillclimbing.}.  To learn more about GAs, please
see~\cite{goldberg1989genetic}.

\begin{table}[h]
    \centering
    \caption{GA Parameters}
    \label{table:gaParms}
    \begin{tabular}{p{6cm}p{6cm}}
    \toprule
    Parameter & Value\\
    \midrule
    Population-Size & 50 (trial \#1), 100 (trial \#2)\\
    Crossover-Type & Uniform \\
    Mutation-Rate & 2\%\\
    Steady-state vs. generational & Generational \\
    Number of Evaluations before restart & 10,000 \\
    Elitist Selection & Yes - Single best member always survives from
    $generation_t$ to $generation_{t+1}$\\
    \bottomrule
    \end{tabular}
\vspace{0.1in}  

\end{table}

We test the GA with two population sizes (50 \& 100) run for an
equivalent number of function evaluations as all of the previous runs
with hillclimbing (500,000).  Additionally, as with the previous
runs, the GA was restarted after 10,000 evaluations.  In the standard
GA, the initial population is comprised of candidate solutions that
were randomly generated.  In the Deep-Opt-GA, the initial population
of candidate solutions is entirely generated from the back-driven
neural network model.

This approach was tested on the same real-valued constraints problems
described in Section~\ref{realconstraints}.  The results\footnote{To
  be consistent with the rest of the paper, the problem evaluation was
  inverted to make the minimization problem a maximization problem;
  hence, larger scores are better.} are shown in
Table~\ref{table:gaResults}.  Deep-Opt-GA outperformed the random
initialization on all 20 instantiations attempted, for both sets of
trials (with population size 50 and 100).  The population size did not
make a significant difference.  Although not directly relevant to the
effects of Deep-Opt, it is interesting to note how these results
compare to the hillclimbing runs described in
Section~\ref{realconstraints}.  Even the simplest NASH (NASH-1), with
a performance of 5206, did better than the GA models tested; though
the GA models were consistently improved with learning. To summarize
the findings in this section, using models to guide search can help
even in search heuristics that operate from more than a single point
-- those that are population-based, such as genetic algorithms.

\vspace{0.1in}
\begin{table}[h]
    \centering
    \caption{GA performance on Graph-Based Constraints Problem (20 Instantiations)}
    \label{table:gaResults}
    \begin{tabular}{p{6cm}p{3cm}p{3cm}}
    \toprule
    Approach & Performance & Number of Wins\\
    \midrule
    Standard GA (population size 50) & 4817 & 0\\
    Deep-Opt-GA (population size 50) & 5120 & 20\\
    \midrule    
    Standard GA (population size 100) & 4825 & 0 \\
    Deep-Opt-GA (population size 100) & 5131 & 20\\        
    \bottomrule
    \end{tabular}
\vspace{0.1in}  
\end{table}

\section {Discussion and Future Work}
\label{conclusions}

We have presented a novel method to incorporate deep-learning with
stochastic optimization.  It is the next instantiation of intelligent
model-based stochastic optimization and follows in the tradition of
the probabilistic model based optimization approaches from the last
two decades of research.  An important aspect of this work is that
\emph{a priori} information about the problem is minimal in setting
the form of the model.  In this study, two multi-layer feed-forward
networks with 5 and 10 hidden layers were used on all of the problems
with no problem-specific modifications.  With the judicious use of
early-stopping in training, the potential downsides for overtraining
were overcome.

A consideration that we did not discuss in this paper is the speed of
optimization.  Between every hillclimbing run, a network is trained
and sampled; this can be a time consuming procedure on the problems
with the large solution-encoding (\emph{e.g.} the discrete constraint
satisfaction problem).  As yet, we have not made attempts to optimize
this; however, in the future, methods that employ fewer training steps
between hillclimbing runs should be explored.  It is likely that will
achieve similarly positive results.

Many of the advances from the rapidly evolving field of deep learning
can be directly incorporated into this work (such as network
shrinking, rapid training, regularization schemes, etc.), as
continuously training networks is the core of the learning system.
Outside of the deep-learning advancements, three avenues of immediate
interest for future explorations are given below.

First, the overarching goal of this paper was to concretely
demonstrate the integration of neural network learning with
optimization, not to promote this system as finalized optimization
system.  To make a finalized optimization system, further exploration
of the algorithm's robustness and behavior is warranted.  For example,
there are many parameters in this system.  In this study, we have
found that the results are most sensitive to the size of $|S|$ and to
the decision of when to restart training the network from scratch --
this happens when new samples are added to $|S|$ and the network fails
to accommodate them in learning (i.e. the error on the samples does
not reduce).  This may be because of the scaling issues mentioned in
Section \ref{scaling} and/or because the weights of the network may
have grown too large.  The beneficial effects of regularization may
be especially pronounced in this application as networks are
constantly being incrementally trained with changing data. Also, we
have found a class of problems for which we have not observed a
benefit of using the modeling (see Section~\ref{2d}).  Are the
problems too easy or too hard, or is an alternate representation
needed?

Second, an alternative to using the network back-driving technique to
generate candidate initialization points is to simply use the network
as a proxy evaluation function for hillclimbing.  In this approach,
hillclimbing (NASH) is conducted directly on the model's output.
After every perturbation, the new candidate-solution is passed through
the network to measure it's \emph{estimated} performance.  This,
unlike network back-driving, does not take advantage of the fact that
the network is differentiable, and rather only uses it as a proxy for
the real evaluation function.  However, it may reveal parts of the
search space that back-driving may not.

The third, and perhaps the most speculative, direction is to measure
if there are transferable features that are learned between different
instantiations of the same problem.  For example, with respect to the
triangle-covering problem presented in Section~\ref{triangles}, once
we have learned how to evaluate how well a set of triangles reproduces
an image, is learning the evaluations for the next image easier?  One
can imagine that low level primitives of how to draw triangles, if
they are indeed learned by the network, may be reusable.  Even if
there is little transference in the current implementation, exploring
problem transference has enormous potential to make this system
automatically more intelligent with time.

\vspace {0.3in}

\subsubsection*{Acknowledgments}

The author would like to gratefully acknowledge Sergey Ioffe and Rif
A. Saurous for their invaluable comments.

\vspace {0.3in}



\medskip

\small

\bibliographystyle{IEEEtran}
\bibliography{optimizationBib}

\end{document}